\documentclass[letterpaper, 10 pt, conference]{ieeeconf}  % Comment this line out if you need a4paper

\IEEEoverridecommandlockouts                              % This command is only needed if 
\usepackage{xcolor} %for textcolor in notes and correction
\usepackage{graphicx} % for pdf, bitmapped graphics files
\graphicspath{{./figures/}}
\usepackage{amsmath} % assumes amsmath package installed
\usepackage{amssymb}  % assumes amsmath package installed
\usepackage{dsfont}
\usepackage{algpseudocode} %Provides algorithmicx. the best. 
\usepackage{tabulary}
\usepackage{algorithm} %to wrap the pseudo code in numbered algorithm env
\usepackage{subcaption}
\usepackage{wrapfig}
\usepackage{hyperref}
\usepackage{cite}
\usepackage{eso-pic}
\usepackage{cancel} %for strike through
\newcommand{\algo}{SafeAPT} %Safe sim-to-real learning using Diversity
\newcommand{\algofull}{Safety-Aware Policy Transfer}

\newcommand{\ie}{\emph{i.e.}} 
\newcommand{\eg}{\emph{e.g.}} 
\DeclareMathOperator*{\argmax}{argmax\text{ }}

\title{\LARGE \bf SafeAPT: Safe Simulation-to-Real Robot Learning using Diverse Policies Learned in Simulation }
 
\author{Rituraj Kaushik$^*$, Karol Arndt and Ville Kyrki
\thanks{*Corresponding author: {\tt\small rituraj.kaushik@aalto.fi}}
\thanks{All authors are affiliated with Intelligent Robotics Group, Dept. of Electrical Engineering and Automation, Aalto University, Finland.}
\thanks{We acknowledge the computational resources provided by the Aalto Science-IT project.}
\thanks{\hrulefill}
\thanks{\scriptsize Video:\url{http://tiny.cc/safeAPT}}
\thanks{\scriptsize Code: \url{https://github.com/riturajkaushik/SafeAPT}}
}

\begin{document}

\AddToShipoutPicture*{\put(50,740){\parbox[b][\paperheight]{\paperwidth}{%
\vfill
\footnotesize
{This work has been submitted to the IEEE for possible publication. Copyright may be transferred without notice, after which this version may no longer \\be accessible.}
}}}

\maketitle
\thispagestyle{empty}
\pagestyle{empty}

\begin{abstract}

The framework of Simulation-to-real learning, \ie{}, learning policies in simulation and transferring those policies to the real world is one of the most promising approaches towards data-efficient learning in robotics. However, due to the inevitable reality gap between the simulation and the real world, a policy learned in the simulation may not always generate a safe behaviour on the real robot. As a result, during adaptation of the policy in the real world, the robot may damage itself or cause harm to its surroundings. In this work, we introduce a novel learning algorithm called \algo{} that leverages a diverse repertoire of policies evolved in the simulation and transfers the most promising safe policy to the real robot through episodic interaction. To achieve this, \algo{} iteratively learns a probabilistic reward model as well as a safety model using real-world observations combined with simulated experiences as priors. Then, it performs Bayesian optimization on the repertoire with the reward model while maintaining the specified safety constraint using the safety model. \algo{} allows a robot to adapt to a wide range of goals safely with the same repertoire of policies evolved in the simulation. We compare \algo{} with several baselines, both in simulated and real robotic experiments and show that \algo{} finds high-performance policies within a few minutes in the real world while minimizing safety violations during the interactions.
\end{abstract}

\section{Introduction}

% RL not directly applicable in robotics
Reinforcement learning (RL) is a promising direction towards allowing robots to acquire new skills through real-world interaction. Despite impressive results in simulated applications, \eg{}, simulated robots~\cite{heess2017emergence}, the application of RL on physical robots is limited primarily due to the data-inefficiency of these algorithms~\cite{kaushik2018multi, chatzilygeroudis2017black}.

% Data-efficiency through sim-2-real
In recent years, the idea of sim-to-real policy adaptation has become a promising alternative to achieve data-efficiency in robot learning using RL~\cite{kaushik2020data, bousmalis2018using}. In this approach, first, a policy is learned in the simulation. Then, that policy is adapted through real-world interactions to cross the reality-gap, the unmodeled or unknown variations between the simulation and the reality.

\begin{figure}
  \vspace{1.0em}
  \centering
  \includegraphics[width=1.0\linewidth]{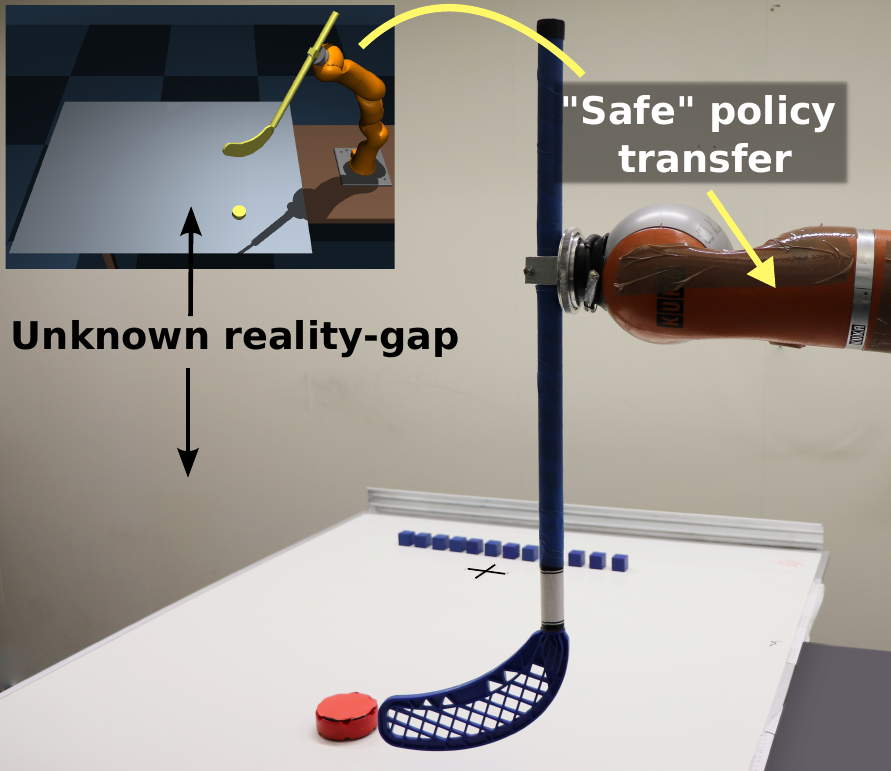}
  \caption{\label{fig:visual_abstract} Safe sim-to-real policy transfer when there is unknown reality-gap (\eg{}, unknown friction) between the simulation and the real world, and when the goal of the task is not specified apriori in the simulation (\eg{}, desired location of the puck is not specified in simulation).}  
  % \vspace{-1.5em}
\end{figure}

\begin{figure*}[ht]
  \centering
  \vspace{1.0em}
  \includegraphics[width=0.9\textwidth]{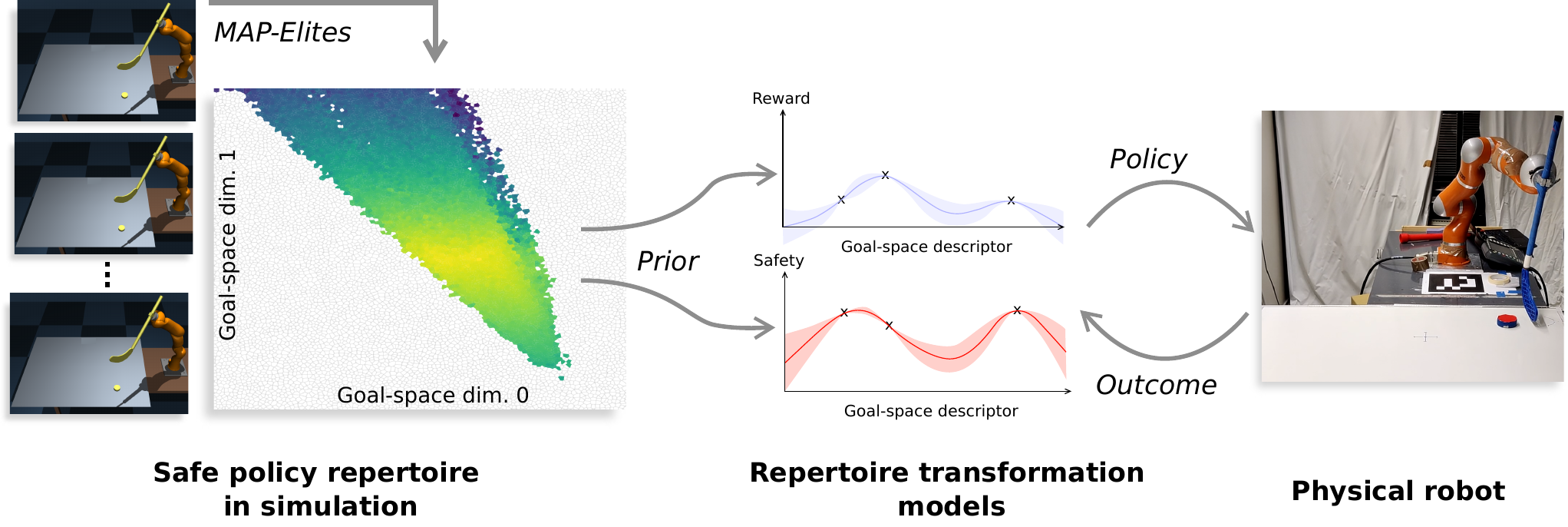}
  \caption{\label{fig:overview} \textbf{Overview:} \algo{} first generates a large set of policies (also called a repertoire) that are maximally safe to achieve diverse goals in various simulated dynamics conditions (\eg{}, friction, joint damage, mass etc.). Then, given an arbitrary goal in the real world, the robot figures out the most suitable policy to maximize the reward by iteratively trying policies from the repertoire. \algo{} ensures data-efficiency and minimizes the safety violations during real-world interactions thanks to the probabilistic reward model and safety model learned online using the simulated data as ``priors''.}
  % \vspace{-1.5em}
\end{figure*}

To improve the data-efficiency further in the sim-to-real policy adaptation approach, repertoire-based learning approaches optimize the policy on a discretized outcome-space, which is often of a lower dimensionality than the policy parameter space~\cite{cully_robots_2015, kaushik2019adaptive}. The outcome-space can be defined as a user-defined space where the outcome or the behaviour of the policies can be specified. For a robot hitting a hockey puck, for instance, the outcome-space can be defined as the 2D space of $\langle x,y \rangle$ coordinate position of the puck after executing the policy on the robot. Similarly, for a walking robot, the outcome-space can correspond to the different types of gaits produced by the policies on the robot. The core idea behind this approach is to evolve a large repertoire (\ie{}, a collection) of high rewarding policies in simulation and associate each of them with a unique outcome in the discretized outcome-space. Then, on the physical robot, the optimal policy is chosen typically through Bayesian optimization~\cite{brochu_tutorial_2010} in the outcome-space. The main hypothesis of this approach is that due to the diversity of the policies in the repertoire, some policies in the repertoire will still produce high rewards on the physical robot even in the presence of a large reality-gap. For instance, a robot with a damaged leg can still walk if the repertoire has a policy to produce a walking gait that does not use the broken leg for walking.   

% Problems in current sim-2-real
Nevertheless, the sim-to-real adaptation approach has one major shortcoming. Due to the reality-gap, there is no certainty that the policy learned in simulation is safe to execute on the physical robot for policy evaluation. The execution of an unsafe policy may damage the robot or its surroundings during the learning process.

% What we propose
In this work, we propose a repertoire-based multi-goal learning approach called \algo{} (\algofull{}) that allows a robot to learn new skills in simulation, and then transfer them safely to the real world. In this approach, first, we evolve a large repertoire of policies to achieve a diverse set of goals in the simulation.
% using an evolutionary algorithm called MAP-Elites~\cite{vassiliades2017using, mouret_illuminating_2015}.
These policies are evolved in such a way that for each goal, the associated policy performs the task as safely as possible within a distribution of diverse dynamics conditions of the robot. Then, on the physical robot, given a specific goal and the safety limit, \algo{} performs Bayesian optimization (BO) on the policy repertoire to maximize the reward while maintaining the constraint on the safety for each trial. To perform this constrained BO, we introduce a new acquisition function called ``Expected safe improvement'' (ESI). The ESI-BO uses two iteratively learned probabilistic models: the reward and the safety transformation models. These models map the outcomes of the policies in the repertoire to the real-world rewards and safety scores. As each policy in the repertoire is associated with a unique outcome (i.e., the associated goal), the transformation models implicitly map the policies to their reward and safety score. To learn these models in a data-efficient manner, we incorporate the simulated results in the repertoire as priors for the models.

% Main hypothesis
The primary hypothesis in this work is that, due to the reality-gap, a policy repertoire evolved in the simulation undergoes a transformation on the outcome-space for the real robot. As a result, the safety and reward associated with those policies are also transformed. We capture the transformations of the safety and the reward functions with Gaussian process regression models~\cite{rasmussen2006gaussian} using the simulated results as prior mean functions.

% Clearly state the contribution
More concretely, our work has the following contributions:
\begin{enumerate}
    \item Data-efficiency: \algo{} allows a robot to learn policies within a few minutes of interaction using the sim-to-real adaptation approach.
    \item Safety: \algo{} minimizes safety violations during real-world interactions.
    \item Multi-goal: Thanks to the diverse policy repertoire evolved in the simulation, \algo{} does not have to re-optimize a policy in the simulation when the real-world goal or the safety limit change.
\end{enumerate}

We compare \algo{} with three baselines, both in simulated and real-world experiments, and demonstrate that \algo{} finds high-performance policies within less than a minute of real-world interaction while minimizing the safety constraint violations compared to the baselines.

\section{Related work}

Several prior works use probabilistic dynamical models to avoid unsafe behaviour \cite{fisac2018general, hewing2019cautious, zhang2020cautious} during learning through trial-and-error. For instance, in \cite{zhang2020cautious}, the agent first uses a model-based RL approach to learn a probabilistic model to capture uncertainty about transition dynamics and catastrophic states. The model is then used in the real world for predicting and avoiding potentially unsafe states.

Shielding-based safe RL approaches typically use a safety-critic to estimate the safety of an action at the current state of the RL agent~\cite{alshiekh2018safe, bharadhwaj2020conservative}. If any action is predicted to be unsafe, the alternative safe action is executed by the agent. Typically, the Bellman equation is used to update the safety critic with sampled transitions from the current policy. However, while training the safety-critic, the agent may violate the safety constraints. Moreover, these approaches are not data-efficient enough to use on physical robots.  

One class of optimization algorithms that has been successfully applied to robotics is Bayesian optimization (BO) \cite{shahriari2015taking, calandra2014experimental}, particularly due to its ability to optimize black-box objectives which are expensive to evaluate.~\cite{gardner2014bayesian} introduce a general framework to incorporate inequality constraints in Bayesian optimization. Similarly,~\cite{sui2015safe, berkenkamp2021bayesian} propose safe Bayesian optimization in the context of parameter tuning in robotics. Nevertheless, BO does not scale well with the dimensionality of the parameters. Thus, on physical robots, BO is practically limited to optimizing around 10 parameters.

% \subsection{Repertoire-based Learning in Robotics}
In order to scale up BO to high-dimensional parameter space, repertoire-based learning in robotics performs the policy optimization on the low dimensional outcome-space. The core idea behind this approach is to first pre-compute a large and diverse set of policies in simulation with a ``quality-diversity'' algorithm \cite{mouret_illuminating_2015, cully2018quality} and associate them with unique low-dimensional discrete outcomes. Then an optimization process (\eg{}, BO) figures out the policy on that discrete outcome space that works best in current dynamics conditions \cite{cully_evolving_2015,cully_robots_2015, duarte2017evolution, sharma2019dynamics}. For instance, IT\&E approach ~\cite{cully_robots_2015} evolves a policy repertoire for a Hexapod robot to walk forward in simulation, but with different walking gaits (here gaits are the outcomes). On the physical robot with a high reality-gap due to a damaged leg, IT\&E performs BO to figure out the gait (and so the associated policy in the repertoire) that makes the robot walk forward.  

To extend IT\&E approach for safety, sIT\&E~\cite{papaspyros2016safety} includes safety constraints as additional dimensions to the policy repertoire. As such, the repertoire now contains diverse policies to perform the same goal oriented task (\eg{}, to walk forward as fast as possible), with different behaviours or outcomes (\eg{}, walking gaits) and with different safety scores. Given the safety constraints, sIT\&E figures out the policy through trial-and-error using constrained BO \cite{gardner2014bayesian}. The main limitation of sIT\&E is that when the goal of the task changes in the real world, the repertoire needs to be evolved again, which is computationally expensive - typically takes several hours. In addition, as the number of safety constrained increases, the dimensionality of the repertoire increases which makes the BO less efficient.   
 
% To the best of our knowledge, none of the prior work on sim-to-real robot learning considers the ``multi-goal'' and ``safety'' criteria together. The work that is the closest to our work is sIT\&E~\cite{papaspyros2016safety}. However, unlike \cite{papaspyros2016safety}, \algo{} explicitly considers diverse dynamics conditions that the robot might face in the real-wold while evolving the repertoire. In addition, unlike sIT\&E, \algo{} is multi-goal and does not require knowledge about the actual real-world goal of the task in simulation.

To the best of our knowledge, none of the prior work on sim-to-real robot learning considers the ``multi-goal'' and ``safety'' criteria together. Unlike prior work, \algo{} explicitly considers diverse dynamics conditions that the robot might face in the real-wold while evolving the repertoire. In addition, \algo{} is multi-goal and does not require the knowledge about the actual goal of the task a priori in simulation.

\section{Problem Statement}

We consider that the dynamics of the robot and its environment can be represented jointly with the following dynamical system:
\begin{align}
  \mathbf{s}_{t+1} &= f(\mathbf{s}_t, \mathbf{a}_t, \boldsymbol{\psi}) + \mathbf{w} \label{eg:system}
\end{align}
where the function $f(\cdot, \cdot, \cdot)$ represents the state transition dynamics, $\mathbf{s}_t$ and  $\mathbf{a}_t$ represent the state of the system and action applied on the system at time $t$, $\mathbf{s}_{t+1}$ is the corresponding next-state of the system, $\boldsymbol{\psi} \in \mathbb{R}^{d_\psi}$ is the dynamics parameter to incorporate different dynamics conditions, and $\mathbf{w}$ is the i.i.d Gaussian noise to account for any unmodeled dynamics and inherent stochasticity of the system. 
% We refer to $\boldsymbol{\psi}$ simply as a ``situation'' that alters the dynamics of the system. 
We assume that the robot (our embodied agent) has access to the function $f(\cdot, \cdot, \cdot)$ in the form of a physics simulator. However, the robot does not know the value of the dynamics parameter $\boldsymbol{\psi}_{real}$ in the real world. Instead, the robot has the knowledge that $\boldsymbol{\psi}_{real} \in \Psi \subseteq \mathbb{R}^{d_\psi}$, where $\Psi$ is the (finite/infinite) set of feasible values of $\boldsymbol{\psi}$ in the real world. 

The task has parametric goals $\mathbf{g} \in \mathbb{G} \subseteq \mathbb{R}^{d_g}$. We assume that the robot is controlled by a deterministic policy (closed/open loop) $\pi_{\boldsymbol{\theta}}$ parameterized by $\boldsymbol{\theta} \in \mathbb{R}^{d_\theta}$ such that $\mathbf{a}_t = \pi_{\boldsymbol{\theta}}(\mathbf{s}_{t}, t)$. Execution of the policy $\pi_{\boldsymbol{\theta}}$ on the system for $N$ steps and for any $\psi$ results in the trajectory $\boldsymbol{\tau} = (\mathbf{s}_{0}, \mathbf{a}_{0}, \mathbf{s}_{1}, \mathbf{a}_{1}, \ldots, \mathbf{s}_{N})$. After execution of the policy, for any given goal $g$, the robot receives a trajectory reward $R(\boldsymbol{\tau}, \mathbf{g})$ and trajectory safety score $C(\boldsymbol{\tau})$. The robot has access to the functions $R(\cdot,\cdot)$ and $C(\cdot)$ to compute the reward and the safety score associated with any trajectory $\tau$.

The robot has to solve the following optimization problem for a specified minimum safety score or ``safety-limit'' $\lambda$ and goal $\mathbf{g}$ through episodic trial-and-error: 
\begin{align}
  \boldsymbol{\theta}^* = &\argmax_{\boldsymbol{\theta}} \mathbb{E}_{\boldsymbol{\tau} \sim \pi_{\boldsymbol{\theta}}} \Big[R(\boldsymbol{\tau}, \mathbf{g})\Big] \label{eq:problem} \\ 
  &\text{subject to } \mathbb{E}_{\boldsymbol{\tau} \sim \pi_{\boldsymbol{\theta}}} \big[C(\boldsymbol{\tau}) \big] \ge \lambda \label{eq:constraint}
\end{align} 

In addition, the constraint \ref{eq:constraint} must be satisfied for any policy evaluation on the physical robot. In other words, we are not simply concerned about the safety of the optimal policy, but we want any policy evaluated on the robot during exploration to also be safe. 

\section{Approach}
Our approach consists of 3 main steps: 

\begin{enumerate}
  \item We generate a repertoire (i.e., an archive or a collection) of policies that produce diverse goal-space outcomes in the simulation. Each policy in the repertoire is as safe as possible on a distribution of simulated dynamics conditions of the real world.
  
  \item Using the simulated results as priors for the Gaussian process models, we learn online how the safety and reward of the policies transformed in the real world.
  
  \item Then, we use BO to evaluate policies on the robot from the repertoire that are safe and potentially improve the reward. We perform steps 2 and 3 iteratively until the task is solved (see Algo.~\ref{algo:repertoire} and~\ref{algo:sim2real}).
\end{enumerate}

These steps are elaborated in the following subsections:

\subsection{Generating the policy repertoire in simulation}
The offline phase of the \algo{} starts with the generation of a policy repertoire. Our objective here is to generate a large set of policies that are as safe as possible in simulation and cover the goal-space $\mathbb{G}$ of the task as widely as possible (i.e., each reachable discretized bin in $\mathbb{G}$ is assigned a policy that maximizes the safety score while reaching goals in the given bin). A policy repertoire $\boldsymbol{\Pi}$ is a set of tuples  $\langle \boldsymbol{\theta}, \boldsymbol{\tau}, \mathbf{x}, c\rangle$, where $\boldsymbol{\theta}$ is the policy parameter, $\mathbf{x} \in \mathbb{R}^{d_g}$ is the goal-space descriptor (\eg{}, Cartesian co-ordinate in a goal-reaching task) associated with the policy, $\boldsymbol{\tau}$ is the trajectory, and $c$ is the safety-score (the higher the better) for the policy.

We uniformly sample $N$ different situations $\boldsymbol{\psi}_{i=0,\ldots, N-1}$ from the set of feasible dynamics conditions $\Psi$. For instance, a dynamics condition may include the mass of the object that the robot is intended to manipulate, the friction in the robot's joints, and so on.

To generate the policy repertoire, we use the quality-diversity algorithm called MAP-Elites~\cite{mouret_illuminating_2015}. MAP-Elites first discretizes the goal-space $\mathbb{G}$ into $K$ cells. Then it randomly initializes $m$ policies $\boldsymbol{\theta}_{i=1:m}$ and evaluates them on each of the dynamics conditions $\boldsymbol{\psi}_{i=0,\ldots, N-1}$ in the simulator. Then it creates the tuples $\langle\boldsymbol{\theta}_i, \boldsymbol{\tau}_i, \mathbf{x}_i, c_i\rangle_{i=1:M}$. Here, $\mathbf{x}_i$ is the mean goal-space descriptor and $c_i$ is the minimum safety score obtained in all the dynamics conditions with the policy $\boldsymbol{\theta}_i$. Then, MAP-Elites attempts to insert the tuples into the respective cells in the repertoire based on their corresponding goal-space outcome. If two tuples fall in the same cell, the tuple with the maximum safety score is inserted. After this initialization, MAP-Elites performs the following steps iteratively until the policy evaluation budget is reached:

\begin{enumerate}
    \item Randomly picks a tuple $\langle\boldsymbol{\theta}_i, \boldsymbol{\tau}_i, \mathbf{x}_i, c_i\rangle$ from the repertoire, and adds a small random variation to the policy $\boldsymbol{\theta}_i$.
    \item Evaluates the policy on all the dynamics conditions to create a new tuple.
    \item Inserts the new tuple into the repertoire if the cell is empty, or, replaces an existing tuple by the new tuple with a higher safety score (discards the new tuple otherwise).
\end{enumerate}

After repeatedly performing the above steps for a sufficient number of times, the repertoire will contain policies that are maximally safe in the simulation over the distribution of the feasible dynamics conditions.
\begin{algorithm}[H]
  \scriptsize
  \caption{Generate safety repertoire}
  \label{algo:repertoire}
  \begin{algorithmic}
    \Require $\mathbb{G} \subseteq \mathbb{R}^{d_g}$ \Comment{Goal-space}
    \Require $\Theta \subseteq \mathbb{R}^{d_\theta}$ \Comment{Policy space}
    \Require $\Psi$ \Comment{Set of feasible real world dynamics conditions}
    \Require $f_{sim}(\cdot, \cdot, \cdot)$ \Comment{The simulator}
    \Require $C(\cdot)$ \Comment{Trajectory safety-score function}
    \Require $R(\cdot, \cdot)$ \Comment{Trajectory reward function}
    \Require $N_{max}$ \Comment{Max. number of evaluation}
    \Require $K$ \Comment{Number of cells in the repertoire}
    % \Require $P(\psi)$ \Comment{Distribution of situations}
    \State
    \Function{Eval}{$\mathbf{\theta}$} \Comment{Policy evaluation function for MAP-Elites}
      \State $\mathcal{D}_{c} \gets \{\}$ \Comment{Empty set of safety scores}
      \State $\mathcal{D}_{x} \gets \{\}$ \Comment{Empty set of goal-space outcomes}
      \For {$\boldsymbol{\psi} \gets \boldsymbol{\psi}_0$ to $\boldsymbol{\psi}_{n-1}$}
        \State $\boldsymbol{\tau}, \mathbf{x} \gets \text{Execute }\boldsymbol{\theta} \text{ on } f_{sim}(\cdot, \cdot, \boldsymbol{\psi})$ \Comment{Obtain the trajectory and outcome}
        \State $\mathcal{D}_{c} \cup \{C(\tau)\}$ 
        \State $\mathcal{D}_{x} \cup \{\mathbf{x}\}$
      \EndFor
      \State $\text{fitness } \gets \text{minimum} (\mathcal{D}_{c})$ 
      \State $\text{descriptor } \gets \text{avegage}(\mathcal{D}_{x})$
      \State \Return fitness, descriptor
    \EndFunction
    \State
    \Function{Repertoire}{\text{}}
      \State $\boldsymbol{\psi}_0, \boldsymbol{\psi}_1, \ldots, \boldsymbol{\psi}_{n-1} \sim U(\Psi)$ \Comment{Uniformly sample n dynamics conditions}
      \State $\boldsymbol{\Pi} \gets $ \textsc{map\_elites} $\Big(f_{sim}(\cdot, \cdot, \cdot),\mathbb{G}, \Theta,\boldsymbol{\psi}_{i=0:n-1},\textsc{Eval}(\cdot),N_{max}, K\Big)$
      \State \Return $\boldsymbol{\Pi}$
    \EndFunction
    \State
  \end{algorithmic}
\end{algorithm}
\subsection{Learning of the reward and the safety model}

In the real world, given a goal $\mathbf{g}$, we assign rewards to the tuples $\langle\boldsymbol{\theta}_i, \boldsymbol{\tau}_i, \mathbf{x}_i, c_i\rangle$ in the repertoire using the trajectory reward function: $r_i = R(\boldsymbol{\tau}_i, \mathbf{g})$. These rewards are inserted into the respective tuples in the repertoire:  $\langle\boldsymbol{\theta}_i, \boldsymbol{\tau}_i, \mathbf{x}_i, c_i, r_i\rangle$. 

We initialize two GP regression models that are used to learn a safety transformation function and reward transformation function in the goal-space $T_{c}: \mathbb{G} \mapsto \mathbb{R}$ and $T_{r}: \mathbb{G} \mapsto \mathbb{R}$. A GP model can be fully defined by the mean function $M(\cdot)$ and covariance function $k(\cdot, \cdot)$:
\begin{align}
  & T_{c}(\cdot) \sim \mathcal{GP}\big(M_{c}(\cdot), k_{c}(x, x') \big) \\
  & T_{r}(\cdot) \sim \mathcal{GP}\big(M_{r}(\cdot), k_{r}(x, x') \big)
\end{align}

If $\mathcal{D}_{c_{1:t}}$ and $\mathcal{D}_{r_{1:t}}$ are the safety and reward observations in the real world for $t$ policies from the repertoire, then the GPs can be calculated as:
\begin{align}
  & P(T_{c}(\mathbf{x})|\mathcal{D}_{c_{1:t}}) = \mathcal{N}\big(\mu_{c}(x), \sigma^2_{c}(\mathbf{x})\big) \label{eq_safety_model}\\
  & P(T_{r}(\mathbf{x})|\mathcal{D}_{r_{1:t}}) = \mathcal{N}\big(\mu_{r}(x), \sigma^2_{r}(\mathbf{x})\big) \text{ where,}\label{eq_reward_model}
\end{align}
\begin{align}
  \mu_{c}(\mathbf{x}) =  & M_{c}(\mathbf{x}) + \boldsymbol{k}_{c}^T(\mathbf{K}_{c}+\sigma^2_{n_{c}}I)^{-1} (\mathcal{D}_{c_{1:t}} - M_{c}(\mathbf{x})) \\
  \mu_{r}(\mathbf{x}) =  & M_{r}(\mathbf{x}) + \boldsymbol{k}_{r}^T(\mathbf{K}_{r}+\sigma^2_{n_{r}}I)^{-1} (\mathcal{D}_{r_{1:t}} - M_{r}(\mathbf{x})) \\
  \sigma^2_{c}(\mathbf{x}) = & k_{c}(\mathbf{x},\mathbf{x}) - \boldsymbol{k}_{c}^T(\mathbf{K}_{c}+\sigma^2_{n_{c}}I)\boldsymbol{k}_{c} \\
  \sigma^2_{r}(\mathbf{x}) = & k_{r}(\mathbf{x},\mathbf{x}) - \boldsymbol{k}_{r}^T(\mathbf{K}_{r}+\sigma^2_{n_{r}}I)\boldsymbol{k}_{r} 
\end{align}
$M_c(\cdot)$ and $M_r(\cdot)$ are prior mean-functions for safety and reward transformation models respectively. For any goal-space outcome $x_i$ in the repertoire, $M_{c}(x_i) = c_i$ and $M_{r}(x_i) = r_i$, $\sigma^2_{n_{c}}$ and $\sigma^2_{n_{r}}$ are the prior noise for the GP models, $\mathbf{K}_{c}$ and $\mathbf{K}_{r}$ are the kernel matrices, $\boldsymbol{k}_{c}$ and $\boldsymbol{k}_{r}$ are the rows of the kernel matrices associated with the query $\mathbf{x}$.

Equations \ref{eq_safety_model} and \ref{eq_reward_model} model how the safety and the reward are transformed in the real world compared to the values stored repertoire. For any policy $\theta_i$ in the repertoire, the safety score and the reward can be predicted using the associated goal-space outcome $x_i$ using equations \ref{eq_safety_model} and \ref{eq_reward_model}.

\subsection{Sim-to-real policy transfer using Bayesian optimization}

We modify the expected improvement (EI) acquisition function~\cite{brochu_tutorial_2010} of BO to filter out the policies in the repertoire that are potentially unsafe to execute on the robot. More concretely, we define a new acquisition function called Expected Safe Improvement (ESI) as follows:   
\begin{align}
   & ESI(\mathbf{x}) = EI(\mathbf{x})\times\mathds{1}_{\lambda}(\mathbf{x}) \label{eq:esi}\\
   & \text{where, } \nonumber \\ 
   & \mathds{1}_{\lambda}(x) = \begin{cases}
                                \label{eq:indicator}
                                0 &\quad\text{if } LCB_{c}(\mathbf{x}) < \lambda\\
                                1 &\quad\text{otherwise } \\ 
                                \end{cases}
\end{align}
$LCB(\mathbf{x})$ is the lower confidence bound on the predicted safety for the policy corresponding to the goal-space outcome $\mathbf{x}$ in the repertoire:
\begin{align}
    & LCB(\mathbf{x}) = \mu_{c}(\mathbf{x}) - \kappa \sigma_{c}(\mathbf{x}) \text{, } \kappa \in \mathbb{R}^+ \label{eq:lcb}
\end{align}
In each episode, a new policy $\boldsymbol{\theta}^+$ is selected from the repertoire by maximizing $ESI(\mathbf{x})$:
\begin{align}
    & \boldsymbol{\theta}^+ \Leftrightarrow \mathbf{x}^+ = \argmax_{\mathbf{x} \in \boldsymbol{\Pi}} ESI(\mathbf{x})
\end{align}
After each episode, the GP models are updated with the new observations (Eq. \ref{eq_safety_model} \& \ref{eq_reward_model}). The process continues until the maximum number of trials is reached.
\begin{algorithm}[H]
  \scriptsize
  \caption{Sim-to-real safe policy transfer}
  \label{algo:sim2real}
  \begin{algorithmic}
    \Require The repertoire $\boldsymbol{\Pi}= \{\langle\boldsymbol{\theta}_i, \boldsymbol{\tau}_i, \mathbf{x}_i, c_i\rangle| i=1:N\}$ \Comment{See Algorithm \ref{algo:repertoire}}
    \Require The goal $\mathbf{g} \in \mathbb{G}$
    \Require Trajectory reward function $R(\cdot, \mathbf{g})$
    \Require Trajectory safety-score function $C(\cdot)$
    \State 
    \ForAll{tuple in $\boldsymbol{\Pi}$}
      \State $r_i \gets R(\boldsymbol{\tau}_i, \mathbf{g})$ \Comment{Compute rewards associated with each tuple}
      \State Insert $r_i$ in the tuple:  $\langle\boldsymbol{\theta}_i, \boldsymbol{\tau}_i, \mathbf{x}_i, c_i, r_i\rangle$ \Comment{Update the repertoire}
    \EndFor
    \State $\mathcal{D} \gets \{\}$ \Comment{Empty dataset}
    \State Initialize GP models $T_{c}(\cdot)$ and $T_{r}(\cdot)$
    \For {$i=1$ to $\textsc{max\_trials}$}
      \State Compute $ESI(\mathbf{x})$ for all tuples in $\boldsymbol{\Pi}$ \Comment{See Eq. \ref{eq:esi} --\ref{eq:lcb}}
      \State $\boldsymbol{\theta}^+ \Leftrightarrow \mathbf{x}^+ = \argmax_{\mathbf{x} \in \boldsymbol{\Pi}} ESI(\mathbf{x})$
      \State $r^{+}, c^{+} \gets Execute(\boldsymbol{\theta})$ \Comment{Observed safety and reward}
      \State $\mathcal{D} \gets \mathcal{D} \cup \{(\mathbf{x}^{+}, r^{+}, c^{+})\}$
      \State Update $T_{c}(\cdot)$ and $T_{r}(\cdot)$ using $\mathcal{D}$
    \EndFor 
  \end{algorithmic}
\end{algorithm}
\subsection{Probability of safety violation}
For any policy in the repertoire with associated goal-space outcome $\mathbf{x}$, the probability of violation of the safety limit $\lambda$ can be computed using the Gaussian error function $erf(\cdot)$ as
% % Refer notes on error function: https://www.gaussianwaves.com/2012/07/q-function-and-error-functions/ 
\begin{align}
  &Pr(c<\lambda) = \frac{1}{2} + \frac{1}{2} erf(\frac{z}{\sqrt{2}}) \text{~~~~where,} \label{eq:prob_of_failure}
  % &\text{Where,} \nonumber \\
  % &z = \frac{\lambda - \mu_{safe}(x)}{\sigma_{safe}(x)} \nonumber \\
  % \implies &\lambda = z\sigma_{safe}(x) + \mu_{safe}(x) \label{eq:z_vale}
\end{align}
\begin{align}
  &z = \frac{\lambda - \mu_{c}(\mathbf{x})}{\sigma_{c}(\mathbf{x})} \\
  \implies &\lambda = \mu_{c}(\mathbf{x})+z\sigma_{c}(\mathbf{x}) \label{eq:z_vale}
\end{align}
Now using the $ESI(\cdot)$ acquisition function (Eq. \ref{eq:esi} \& \ref{eq:indicator}), BO only considers policy to test on the real robot that have LCB on safety at least equal to $\lambda$, i.e., 
\begin{align}
  \lambda \le \mu_{c}(\mathbf{x}) - \kappa \sigma_{c}(\mathbf{x}) \text{ (from Eq. \ref{eq:indicator} \& \ref{eq:lcb})} \label{eq:lcb2}
\end{align}
Now using Eq. \ref{eq:z_vale} in \ref{eq:lcb2}
\begin{align}
  z \le -\kappa \label{eq:bound_on_z} 
\end{align}

Since $erf(\cdot)$ is a monotonically non-decreasing function of $z$, using \ref{eq:bound_on_z} in \ref{eq:prob_of_failure}:
\begin{align}
  &Pr(c<\lambda) \le \frac{1}{2} + \frac{1}{2} erf(\frac{-\kappa}{\sqrt{2}}) \label{eq:upperbound_failure}
\end{align}
The inequality in \ref{eq:upperbound_failure} is the upper bound on the safety violation assuming that the GP accurately captures the mean and variance of the safety score associated with a policy in the repertoire. From \ref{eq:upperbound_failure} we see that a higher $\kappa$ value lowers the upper bound on the probability of violating the safety limit. Intuitively, a higher $\kappa$ value means that we are less certain about the mean prediction of the safety. Thus, BO selects policies that have mean safety prediction well above the specified safety limit, making failure probability lower. However, setting a very high value of $\kappa$ will restrict BO from testing policies that are slightly ``risky'' but can potentially give a higher reward. In other words, $\kappa$ sets the trade-off between the safety and reward maximization objectives.

\section{Experimental Setup}
We evaluate \algo{} on three simulated and one real-world tasks, and compared the results with three baselines: 
\begin{enumerate}
  \item CBO: Constrained Bayesian Optimization with learned reward and safety model~\cite{gardner2014bayesian} We expect this baseline to be less data-efficient as the optimization happens directly on the policy parameter space. 
  \item \algo{} (no GP-safety): An ablation baseline of our proposed algorithm, where sim-to-real safety transformation function is not learned from the real-world data. Instead, safety priors stored in the repertoire are assumed to be valid in the real world. This baseline shows the importance of learning the safety model from the real world interaction, even though the repertoire has policies that are potentially ``at least safe'' over a distribution of dynamics conditions.
  \item \algo{} (single dynamics): An ablation baseline of our proposed algorithm, where only one dynamics condition (randomly sampled from $U(\Psi)$) is used to generate the policy repertoire. This baseline evaluates the importance of using multiple dynamics situations to generate the repertoire.
\end{enumerate}
The goal of these experiments is to evaluate \algo{} against the baselines on (1) the data-efficiency, (2) the rate of safety violations during the real-world trials, and (3) the performance (reward) of the final policy.
\begin{figure*}[ht!]
  \vspace{1.0em}
  \centering
  \includegraphics[width=0.8\linewidth]{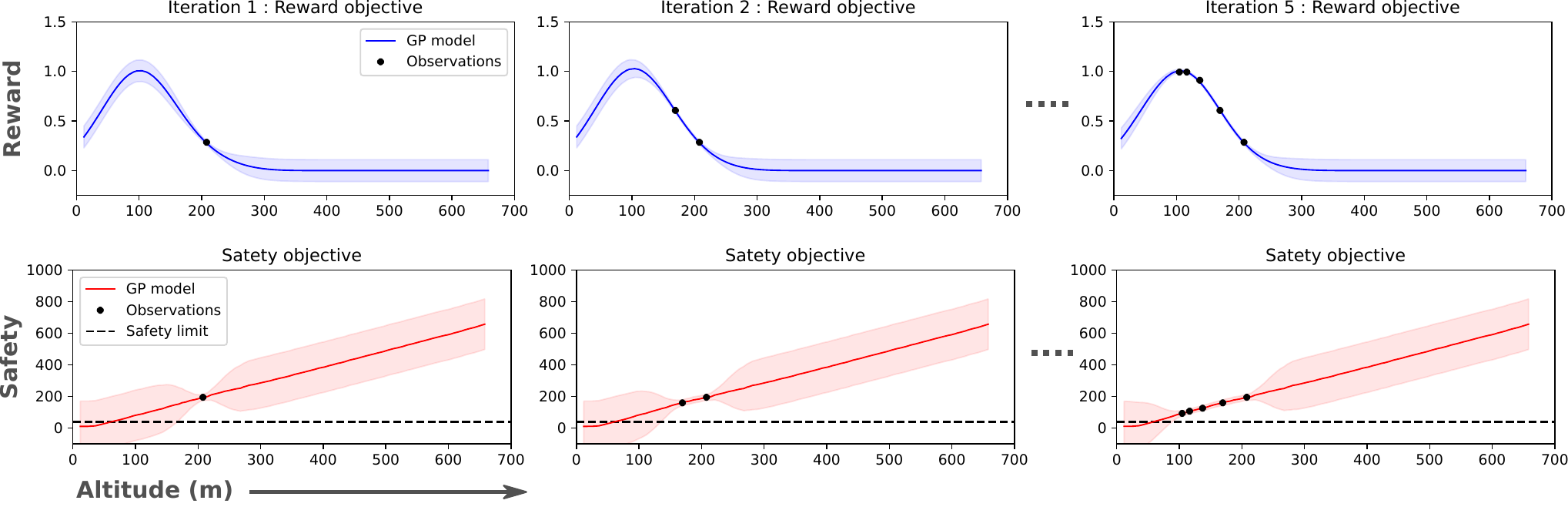}
  \caption{\label{fig:asteroidlanding_optimization} \textbf{GP updates in Asteroid landing experiment: } Plots show how reward and safety GP models are updated using the observations after each episode and how \algo{} cautiously improves the reward while maintaining safety at each trial.} 
  % \vspace{-1.5em}
\end{figure*}

\begin{figure}[ht]
  \centering
  \includegraphics[width=1.0\linewidth]{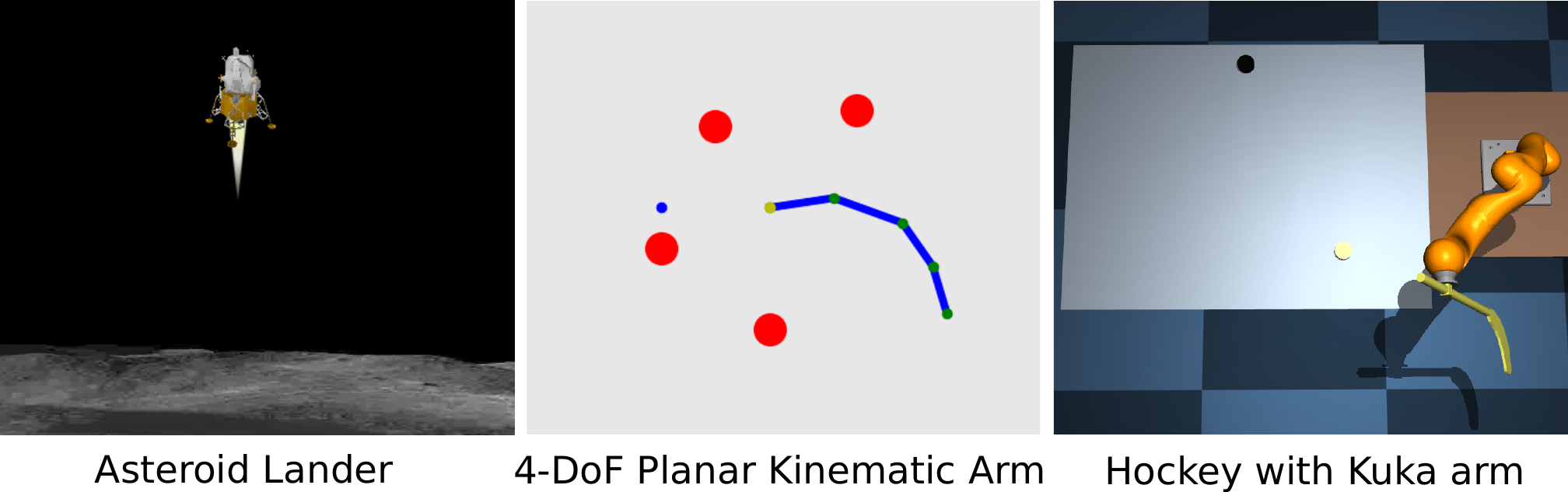}
  % \vspace{-1.5em}
  \caption{\label{fig:simulated_experiments} \textbf{Simulated experiments}} 
  % \vspace{-1.5em}
\end{figure}

\begin{figure*}[ht!]
  \vspace{1.0em}
  \begin{subfigure}{1.0\textwidth}
    \centering
    \includegraphics[width=0.8\linewidth]{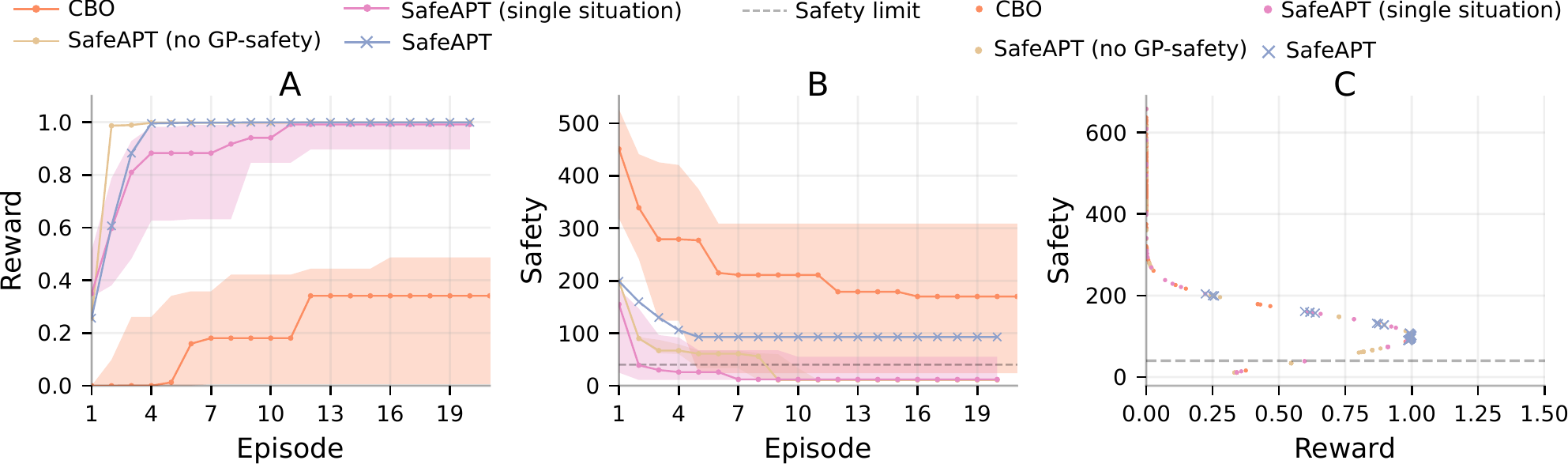}
    \caption{\label{fig:asteroidlanding_episodic} \textbf{Asteroid landing experiment}}  
    \vspace{1.5em}
  \end{subfigure}
  
  \begin{subfigure}{1.0\textwidth}
    \centering
    \includegraphics[width=0.8\linewidth]{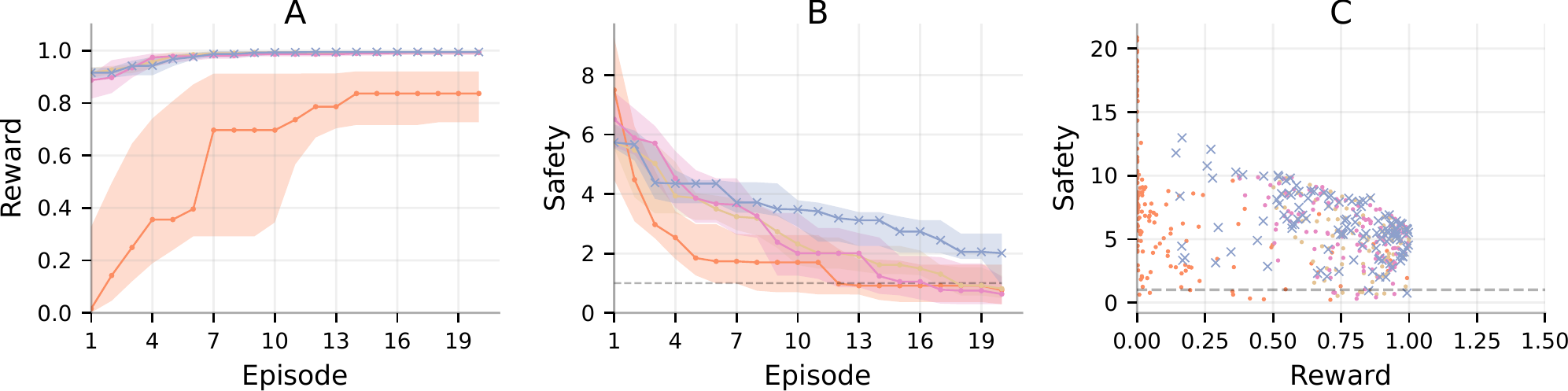}
    \caption{\label{fig:kinematicarm_episodic} \textbf{Kinematic arm experiment}}
    \vspace{1.5em}
  \end{subfigure}
  
  \begin{subfigure}{1.0\textwidth}
    \centering
    \includegraphics[width=0.8\linewidth]{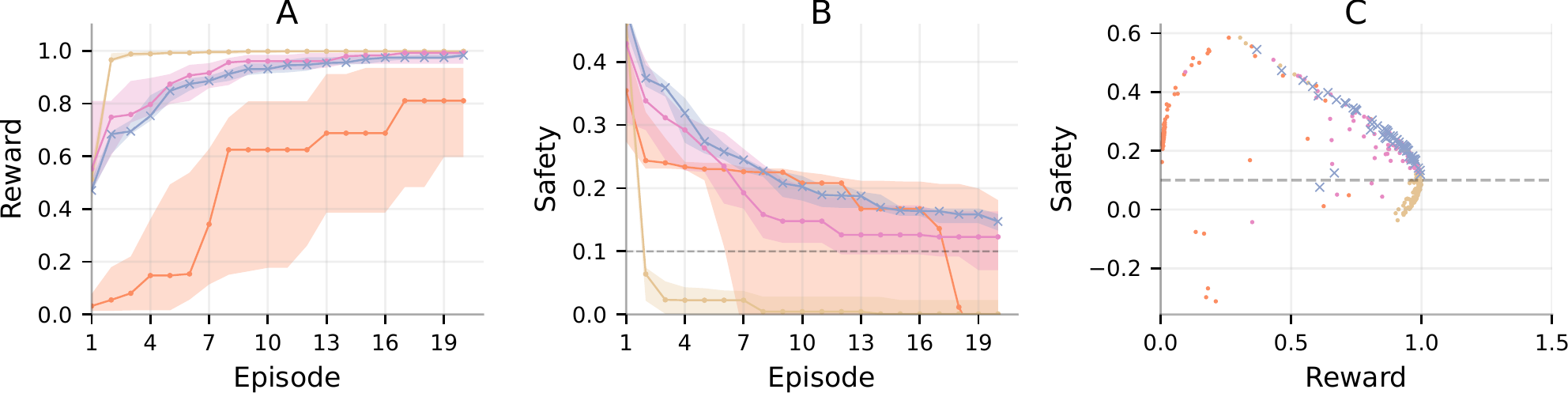}
    \caption{\label{fig:hockeypuck_episodic} \textbf{Hockey-puck experiment with a Kuka arm}}  
  \end{subfigure}
  \caption{\label{fig:exp_plots} For the experiments (a),(b), and (c) the plots A and B show the medians, 25 and 75 percentiles of the reward (plot A) and safety score (plot B) per episode for 15 replicates. Plot C shows the distribution of the executed policies on the reward-safety space. From the plots it can be observed that \algo{} finds higher rewards while staying above the specified safety limit. Additionally, the baselines violate the safety more often than our approach.}
  \end{figure*}
\subsection{Asteroid landing task}
In this task, a simulated asteroid lander has to identify the parameters of a policy that takes it to a given altitude of 100 meters and hovers there. While learning the policy, the lander should not go below a safe altitude of 40 meters. The gravity can vary from $3 m/s^2$ to $10 m/s^2$, and the true value of gravity is unknown to the algorithms during the test time.

Here, the policy is a PID velocity controller, whose three coefficients as well as a sequence of five vertical velocity set-points collectively form the policy parameters ($8D$ policy space). The duration of each episode is $15$ seconds. The goal-space descriptor is the $1D$ altitude of the lander. The trajectory reward is inversely proportional to the distance between the desired altitude and the final altitude achieved after the execution of the policy on the lander. The safety score is the minimum altitude encountered in the trajectory.

\subsection{Planar-arm goal reaching task}

In this task, a 4-DoF planar kinematic arm (shown in the middle of Figure~\ref{fig:simulated_experiments}) has to reach a specified goal 2D location (marked in blue), avoiding 4 unsafe regions (marked in red). During learning, the end-effector should maintain at least 1 unit distance from the unsafe regions. The link lengths of the arm can vary between 4 and 7 units in the simulation. During testing, the algorithms do not have any information about the true link lengths of the arm.

Here, the policy is a feed-forward neural network with $204$ parameters that takes in the current joint angles as input and outputs joint velocity commands at every time-step. The episode length is $50$ timesteps ($5$ seconds). The goal-space descriptor is the 2D coordinate space where the goal lies (scaled to $[0,1]^2$). The trajectory reward is the sum of the rewards collected along the trajectory. At any time-step, the reward is inversely proportional to the distance to the goal from the end-effector of the arm. The safety score is defined as the closest distance to the unsafe regions from the end-effector encountered in the trajectory.

\subsection{Kuka-arm hockey task}
\begin{table}
  \footnotesize
  \renewcommand{\arraystretch}{1.2} %vertical padding
  \begin{tabular}{ m{8em} | m{4.9em} | m{4.9em} | m{6em}}
      \hline
       & Asteroid lander & Kinematic arm & Hockey task \\
      \hline
      Ours & $\mathbf{0.0 \pm 0.0}$ & $\mathbf{0.33 \pm 0.47}$ & $\mathbf{0.07 \pm 0.25}$ \\
      \hline
      Ours\newline(no GP-safety)  & $2.20 \pm 3.06$ & $0.93 \pm 0.85$ & $15.73 \pm 3.84$ \\
      \hline
      Ours\newline(single dynamics)  & $2.0 \pm 2.03$ & $1.13 \pm 1.20$ & $0.73 \pm 1.24$ \\
      \hline
      CBO  & $1.13 \pm 2.36$ & $0.93 \pm 0.85$ & $1.27 \pm 1.30$ \\
      \hline
  \end{tabular}
  \caption[Safety violation]{\small \label{tab:safety_violations} \textbf{Safety violations per experiment:} the number of times the safety limit is violated in each experiment of 20 episodes (mean and standard deviation over 15 replicates).}
  \vspace{-2em}
  \end{table}
This task involves a Kuka LWR 4+ robot arm hitting a hockey puck with a stick such that the puck slides to the desired target position, following~\cite{arndt2019meta, arndt2021fewshot}. During learning, the puck should be at least $0.1$ meters away from the edge of the table (the safety constraint). We vary the value of the friction coefficient between the puck and the sliding surface between $0.4$ to $0.7$ in simulation during training. The algorithm does not have any information about the true value of the friction. The simulation is done using MuJoCo~\cite{Todorov2012MuJoCoAP}; the simulated setup is shown in Figure~\ref{fig:simulated_experiments}.

Here, we use an open-loop policy, which is basically the decoder stage of a denoising auto-encoder. It decodes a $119$ dimensional vector (i.e., the policy parameter $\theta$) to a $119$ dimensional joint position trajectory vector. The decoder was trained beforehand on a dataset of the joint position trajectories that produce different striking motion on the robot. The time required for each episode is $9$ seconds. The goal-space descriptor here is the 2D coordinate space on the table where the puck should land (scaled to $[0,1]^2$). The reward is inversely proportional to the distance between the puck and the goal location. The safety score is defined as the minimum distance from the puck to the edges of the table.

We also built a real-world version of the hockey-puck setup. The setup comprises of a Kuka LWR4+ arm (the same as was used in simulation) equipped with a plastic floorball stick. For the experiment, we used an ice hockey puck, with a whiteboard as a low-friction sliding surface.
The position of the puck is measured by a ceiling-mounted Kinect camera. The safety area is demarcated by a row of wooden cubes. The target position was placed $10$cm away from the safety boundary. For this experiment, we use the same repertoires that were used in simulated experiments. The real-world setup is visualized in Figure~\ref{fig:visual_abstract}.

For all the experiments, we used squared exponential kernels~\cite{rasmussen2006gaussian} for the GP models. The hyperparameters of the GPs and BO are tuned (through grid search) in simulation by evaluating \algo{}'s performance on ``simulation-to-simulation'' policy transfer with different dynamics conditions.  

\section{Results and Discussion}

% In the Asteroid landing task, since the goal-space dimension is 1D, we can easily have an intuition about how the \algo{} works. 
To provide some intuition behind the adaptation process with \algo{}, Figure~\ref{fig:asteroidlanding_optimization} shows the learning of the GP transformation models for the safety and the reward function in the Asteroid landing task. Thanks to the repertoire generated in the diverse simulated conditions, these GP models start with priors, which help to learn faster with only a few data points from the real world. As it can be seen, in the first trial, the lander successfully hovers around 200 meters from the surface considering the high uncertainty about the safety of the policies below that level. The lander then cautiously tries policies from the repertoire that potentially improve the reward without violating the safety limit until it finds the policy that hovers the lander at an altitude of $100$ meters.

From the plots (Figure \ref{fig:exp_plots}) for the simulated experiments, we observed that \algo{} finds at least as high rewarding policies as the baselines while maintaining the safety constraint throughout the whole adaptation process. On the other hand, the baselines violate the safety constraints during trial-and-error learning more frequently than \algo{} (Table \ref{tab:safety_violations}). In all the experiments, due to the lack of the learned safety model, the baseline \algo{} (no GP-safety) maximizes the reward greedily and is unable to maintain the safety constraint. The baseline \algo{} (single dynamics) performs better than \algo{} (no GP-safety) in maintaining the safety due to the learned safety model. However, it performs worse than \algo{} due to the lack of diversity of dynamics in the simulations. As expected, due to the policy optimization directly on the high dimensional parameter space, CBO is not able to compete with repertoire-based counterparts in terms of reward maximization. In terms of safety, CBO performs only slightly better than the other baselines.

\begin{table}
  \vspace{1.0em}
  \centering
  \footnotesize
  \renewcommand{\arraystretch}{1.5} %vertical padding
  \begin{tabular}{ m{8em} | m{8em} | m{7.3em}}
      % \hline
       & \textbf{Maximum reward before violation} &  \textbf{Violations per experiment} \\
      \hline
      \algo{} & $0.97 \pm 0.04$ &  $0.0 \pm 0.0$ \\
      \hline
      \algo{}\newline(no GP-safety)  & $0.89 \pm 0.15$ & $4.4 \pm 4.4$ \\
      \hline
      \algo{}\newline(single dynamics)  & $0.79 \pm 0.31$ &  $3.0 \pm 1.6$ \\
      \hline
      CBO  & $0.43 \pm 0.35$ & $0.72 \pm 0.86$ \\
      \hline
  \end{tabular}
  \caption[Results of physical Kuka-arm hockey task]{\small \label{tab:real_kuka_exp} \textbf{Results of physical Kuka-arm hockey task:} Tables shows that on average, \algo{} achieves higher reward before any safety violation compared to the baselines (8 replicates, each with budget of 15 episodes)}
  \vspace{-2em}
  \end{table}

%NOTE script: plot_real_kuka.py
In the physical Kuka hockey task, \algo{} achieves not only higher reward (out of the maximum possible reward of $1$) but also complies with the safety constraint (no safety violations in 8 replicates with independently generated repertoires; see Table \ref{tab:real_kuka_exp}). Like in previous experiments, the baseline \algo{} (no GP-safety) violates the safety constraint more frequently than the others. Contrary to \algo{} (no GP-safety), \algo{} (single dynamics) shows fewer safety violations due to the presence of the learned safety model. However, due to the lack of diverse dynamics conditions in simulation, it fails to achieve as good reward as \algo{} (no GP-safety) in the real world. As expected, due to the high dimensional policy parameter space, CBO fails to achieve as high rewards as the repertoire-based counterparts. Nevertheless, thanks to the constraints in the Bayesian optimization, CBO violates the safety constraints less frequently than the other baselines.

To summarize, both the simulated and physical experiments confirm that due to the lack of prior knowledge derived from diverse simulated situations, CBO fails in achieving satisfactory rewards and maintaining safety constraints. The ablation baselines \algo{} (no GP safety) and \algo{} (single dynamics) confirm that the diversity in the simulated conditions and learning of the safety transformation model help \algo{} to not only achieve higher reward in a data-efficient manner but also to maintain the safety constraint during learning in the real world.

\section{Conclusion}
Learning new skills in a data-efficient manner through real-world interaction is an open problem in robotics. The problem becomes even more challenging when a robot must ensure safety during interaction in the real world while learning a new skill. In this paper, we proposed a sim-to-real multi-goal learning algorithm called \algo{} for safe robot learning in the real world. \algo{} inherits the typical limitation of repertoire based learning, i.e., the pre-computed policies can be sub-optimal if the discretization of the goal space is not dense enough. Nevertheless, if further policy refinement is desired after performing \algo{}, the fine-tuning of the policy can be performed on the parameter space safely using algorithms like Safe-Opt~\cite{berkenkamp2021bayesian}. We believe that sim-to-real learning approaches like \algo{} can be useful in robot-learning applications where a small mistake by the robot can incur a high cost or a complete failure of the mission, \eg{}, in space or deep-sea applications.

% \clearpage
\bibliographystyle{IEEEtran}
\bibliography{mybib}

\end{document}